\documentclass{article}

\usepackage{PRIMEarxiv}

\usepackage[utf8]{inputenc} 
\usepackage[T1]{fontenc}    
\usepackage{hyperref}       
\usepackage{url}            
\usepackage{booktabs}       
\usepackage{amsfonts}       
\usepackage{nicefrac}       
\usepackage{microtype}      
\usepackage{lipsum}
\usepackage{fancyhdr}       
\usepackage{graphicx}       
\usepackage{subfig}         
\graphicspath{{media/}}     
\usepackage{comment}
\usepackage{xcolor}
\usepackage[T1]{fontenc}
\usepackage{rotating}
\usepackage{array}
\usepackage{textcomp}
\usepackage{gensymb}
\usepackage{makecell}

\usepackage{listings}
\title{Machine-interpretable Engineering Design Standards for Valve Specification 
\thanks{\textit{\underline{Citation}}: 
\textbf{Gjerver et al. Machine-interpretable Engineering Design Standards for Valve Specification. Pages.... DOI:000000/11111.}} 
}

\author{
  Anders Gjerver, Rune Frostad, Vedrana Barisic \\
  Aibel AS \\
  Oslo, Norway\\
  \texttt{\{anders.gjerver, rune.frostad, vedrana.barisic\}@aibel.com} \\
   \And
  Melinda Hodkiewicz, Caitlin Woods \\
  University of Western Australia \\
  Perth, Western Australia\\
  \texttt{\{melinda.hodkiewicz, caitlin.woods\}@uwa.edu.au} \\
  \And
  Mihaly Fekete \\
  Equinor \\
  Oslo, Norway \\
  \texttt{MIHF@equinor.com} \\
  \And  
  Arild Braathen Torjusen, Johan W Kluwer\\
  DNV \\
  Oslo, Norway \\
   \texttt{\{arild.braathen.torjusen, johan.wilhelm.kluewer\}@dnv.com} \\
}

\begin{document}
\maketitle

\begin{abstract}

Engineering design processes use technical specifications and must comply with standards. Product specifications, product type data sheets, and design standards are still mainly document-centric despite the ambition to digitalize industrial work. In this paper, we demonstrate how to transform information held in engineering design standards into modular, reusable, machine-interpretable ontologies and use the ontologies in quality assurance of the plant design and equipment selection process.
We use modelling patterns to create modular ontologies for knowledge captured in the text and in frequently referenced tables in International Standards for piping, material and valve design. These modules are exchangeable, as stored in a W3C compliant format, and interoperable as they are aligned with the top-level ontology ISO DIS 23726-3: Industrial Data Ontology (IDO).  

We test these ontologies, created based on international material and piping standards and industry norms, on a valve selection process. Valves are instantiated in semantic asset models as individuals along with a semantic representation of the environmental condition at their location on the asset. We create "functional location tags" as OWL individuals that become instances of OWL class Valve Data Sheet (VDS) specified valves. Similarly we create instances of manufacturer product type. Our approach enables automated validation that a specific VDS is compliant with relevant industry standards. Using semantic reasoning and executable design rules, we also determine whether the product type meets the valve specification. Creation of shared, reusable IDO-based modular ontologies for design standards enables semantic reasoning to be applied to equipment selection processes and demonstrates the potential of this approach for Standards Bodies wanting to transition to digitized Smart Standards. 

\end{abstract}

\keywords{standard, reasoning, semantic model, data modelling, industrial ontology, equipment specification}

\section{Introduction}

Engineering standards contain technical specifications for a product, service, or organisational procedure. Compliance with a standard confirms that a given product, service, or procedure meets a specific quality and/or technical specification. Compliance with standards is required for organisations to access markets, ensure compatibility between physical, digital and cyber-physical products in an ecosystem, and in some cases to meet regulatory requirements \cite{busch2011,gamito2023influence, blind2020drivers}. 

A large body of rules has been developed over many decades to ensure the safe and economical design, fabrication, and testing of equipment, structures and materials for chemical process plants and equipment. Standard development organisations (SDO), professional societies, trade groups, insurance underwriting companies and government agencies have codified these rules in standards.
The focus of this paper is on the engineering design standards published by the American Petroleum Institute (API), American Society of Mechanical Engineers (ASME) and the American Society for Testing Materials (ASTM).  Examples of items standardized in this sector include the Nominal Sizes (NPS/DN), pressure ratings  and wall thickness of piping, specifications on the composition of alloys and testing procedures. Safe design practices set out in many of these Standards are often used in commercial contracts and accepted by insurance underwriters, so they are widely observed \cite{couper2005chemical}.  As an example of the sheer number of standards that guide engineering work, the API (an SDO) has developed more than 1100 standards, guidelines, and other documents for the oil and gas industry, while the ASTM represents the interests of more than 270,000 companies and administers over 12,000 standards for materials \cite{APIreport, ANSI}. 

These design standards are developed and maintained by subject matter experts coordinated by the SDOs. Most published standards are currently available only as paper or PDF-format digital copies. Standards users do not have access to machine-interpretable content.  All standards are proprietary to the SDO that publish them, meaning that the strides we have seen from the use of large language models to help people query and find information cannot be applied to these technical standards. Individual and corporate users of engineering standards who pay substantial sums to access the standards (a single standard can cost over US\$100) cannot copy the standard (or parts thereof) in neither paper nor digital form. Engineering design houses, asset operators and equipment product manufacturers pay substantial annual costs to provide access to these standards for engineers in their own organizations. Engineers have developed proprietary workflows for specific process design steps and products that capture information from these standards. Each new graduate engineer must learn to find the right standard, search for specific content and extract information from numerous standards as part of specifying or qualifying equipment selections. The opportunity for semantic machine-interpretable standards to improve the efficiency and quality of the engineering design process is immense.

\textbf{Our goal is the creation of machine-interpretable data from engineering design standards to automate steps in the design and product specification for Valves in process plant.} 

\section{Background}

\subsection{Engineering Design Standards}

\begin{figure}[h!]
    \centering
    \includegraphics[width=0.9\linewidth]{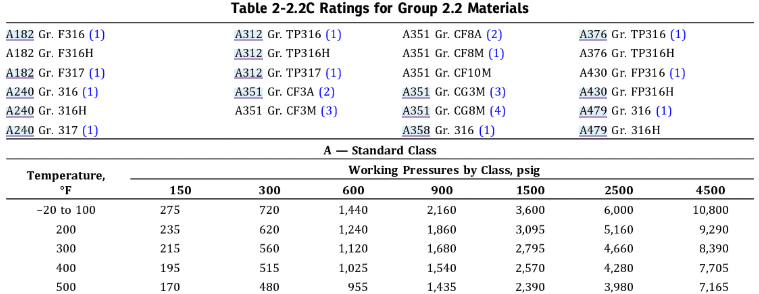}
    \caption{Part of ASME B16.34 Table 2-2.2 showing Group 2.2 materials and acceptable pressure and temperature ranges. Reprinted from ASME B16.34 Valves Flanged, Threaded and Welding End, Edition year 2020 by permission of The American Society of Mechanical Engineers. All rights reserved}
    \label{fig:ASMEB16.34_2.2}
\end{figure}

Figure \ref{fig:ASMEB16.34_2.2} shows an example of data captured in a table in an engineering design standard.  These standards are currently only accessible in paper or pdf formats.  The table is from the ASME B16.34 Valves Flanged, Threaded and Welded End Sets standard. It is typical of an engineering standard  in that it prescribes certain conditions, in this case, acceptable working pressures by pressure rating class (150, 300 etc.) for different temperature ranges for materials classified as belonging to a certain Material Group (Group 2.2). The information in this table can be interpreted as a set of rules. It is these rules and the individual data (the pressures and temperatures, the material specifications, and the numerical ranges) that must be transformed into a machine-interpretable format founded on open and shareable concepts and principles.  

In the domain of plant design the ASME B31 series of design codes list industry standards to which plant objects must comply. Table 1 shows a list of piping and valve specifications that the process industry engineering design community need to be machine-interpretable. Also listed are the number of pages in each standard, the number of tables (part of one table is shown above in Figure \ref{fig:ASMEB16.34_2.2} ) and the number of cross-referenced standards. These documents are highly complex, and their use requires significant training, expertise and concentration by the engineers who are to use them. 

\begin{table}[htbp]
\begin{tabular}{|p{2.4 cm}|p{6.0 cm}|p{1.1 cm}|p{1.3 cm}|p{1.3 cm}|p{1.6 cm}|}
\hline
Standard ID     & Name                                                        & \makecell{Current \\ version} & \# pages & \# tables & \makecell{\# cross- \\ referenced \\ standards} \\ \hline
API 6D          & Pipeline and Piping Valves                                  & 2021            & 184      & 21        & 100                           \\ \hline
ASME B16.5      & Pipe Flanges and Flanged Fittings                           & 2020            & 250      & 143     & 100                           \\ \hline
ASME B16.10     & Face-to-Face and End-to-End Dimensions of Valves            & 2017            & 58       & 10         & 22                            \\ \hline
ASME B16.34     & Valves - Flanged, Threaded and Welded End                   & 2020            & 228      & 106         & 63                            \\ \hline
ASTM A182/A182M & Forged or Rolled Alloy and Stainless Steel Pipe Flanges, Forged Fittings,   and Valves and Parts for High-Temperature Service  & 2024 & 34 & 4 & 96  \\ \hline
ASTM A240/A240M & Chromium and Chromium-Nickel Stainless Steel Plate, Sheet and Strip for   Pressure Vessels and for General Applications        & 2024 & 12 & 3 & 100 \\ \hline
ASTM A351/A351M & Specification for Castings, Austenitic, for Pressure-Containing Parts                                                          & 2024 & 7  & 3 & 35  \\ \hline
ASTM A358/A358M & Electric-Fusion-Welded Austenitic Chromium-Nickel Stainless Steel Pipe   for High-Temperature Service and General Applications & 2024 & 8  & 2  & 33  \\ \hline
ASTM A376/A376M & Seamless Austenitic Steel Pipe for High-Temperature Service & 2022            & 7        & 2         & 11                            \\ \hline
ASTM A430       & Standard Specification for Austenitic Steel Forged and Bored Pipe for High-Temperature Service & 1991 & 5    & 3          & 4                              \\ \hline
ASTM A961/A961M & Common Requirements for Steel Flanges, Forged Fittings Valves and Parts   for Piping Applications                             & 2024 & 9  & 1 & 100 \\ \hline
\end{tabular}
\caption{List of international standards commonly referenced in piping and valve design. The table illustrates the length and complexity of each standard including the number of other standards each cross-references}
\label{tab:standards}
\end{table}

\subsection{Standards Digitization}

Standards digitization refers to the process of creating a new (or transforming an existing)  standard so that the rules and characteristics carried by the standard can be read, transmitted and used through digital devices \cite{liu2021development}. Standards digitization is part of a wider move to industrial digitization which is a ``process that aims to improve an entity by triggering significant changes to its properties through combinations of information, computing, communication, and connectivity technologies" \cite{vial2021understanding}.  
Digitization presents a major change in how we work and do business. The current situation in which data is stored and managed in countless unconnected digital solutions is a barrier to efficient decision making and development of competitive products and services \cite{matt2023industrial}. This problem is particularly acute in the domain of international standards for product lifecycle management (PLM) \cite{rachuri2008information}.

\begin{figure}[ht]
    \centering
    \includegraphics[width=0.8\linewidth]{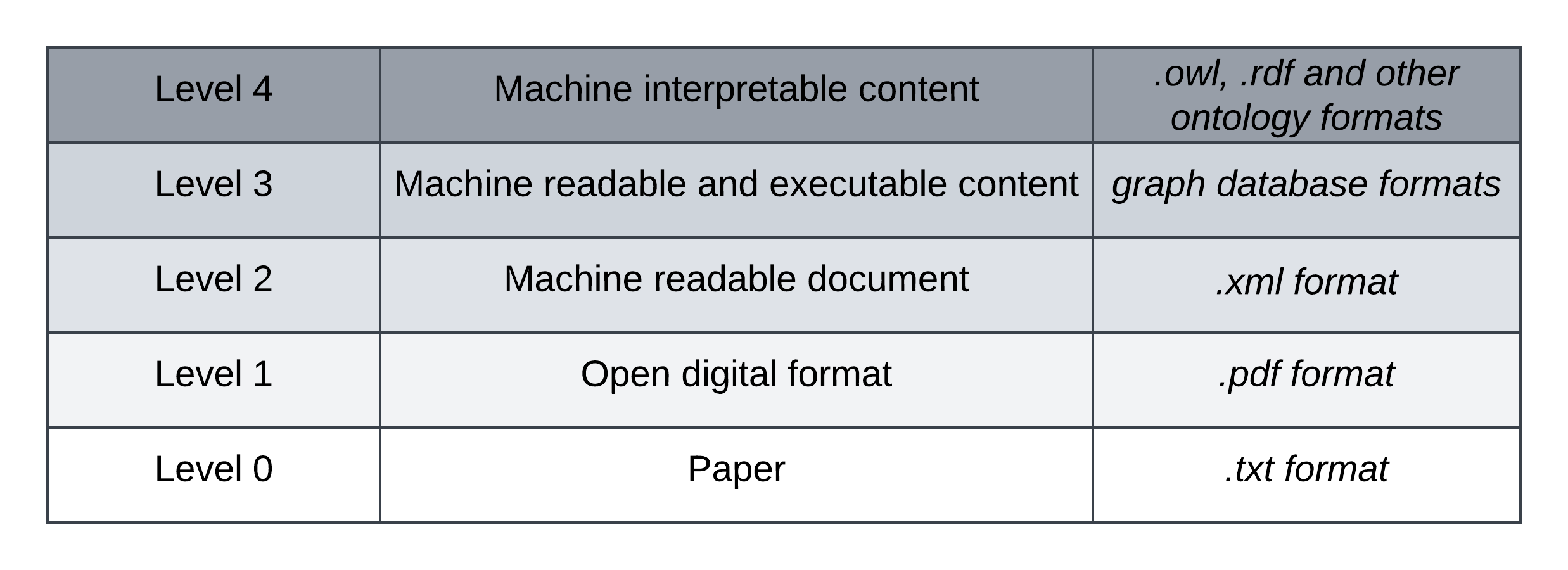}
    \caption{Levels in the ISO/IEC SMART standards hierarchy}
    \label{fig:ISO-IEC}
\end{figure}

Two of the largest global engineering standards organisations ISO and IEC (International Electrotechnical Commission) have prioritised digitalization of standards in their strategies \cite{IECSMARTreport, IECSMART2report}. In 2019 ISO and IEC proposed the SMART concept to move to machine-applicable, readable and transferable standards. ISO/IEC has identified five levels for SMART standards. Standards available on paper (level 0), standards in an open digital pdf format (level 1), machine-readable documents with structured XML content (level 2), machine-readable and executable content (level 3) and machine-interpretable content based on information modelling that expresses content and relations between elements (level 4) \footnote{\url{https://www.iso.org/smart}}. Standards available at levels 3 and 4 are called SMART standards \cite{IECSMART2report} as shown in Figure \ref{fig:ISO-IEC}.

In 2025 ISO/IEC has mandated that all new standards be developed in an open-XML based digital format using their Online Standards Development platform (OSD) \cite{ISO_OSD}. Other SDO are either planning to use OSD or are developing their own standards development platforms \cite{CEN_CENELC}.
While OSD will be a major step forward in providing an XML format and document metadata for new standards there is, as yet, no proposed solution to deal with how to translate the thousands of existing standards into machine-interpretable formats. Important challenges for engineering data include the representation of data in tables and the cross referencing of data between tables in the other standards using a common semantic data model. Specific work is underway to enable the semantic parsing of engineering requirements \cite{holter2020semantic, holter2023reading} as well as to automate extraction of engineering formulas \cite{luttmer2022smart}. In this paper we focus on semantic representation to enable activities at level 4. We address the following goals identified by IEC \cite{IECSMART2report}: 1) standards that can be interpretable by machine without human intervention and, 2) automated compliance check of data to assure conformity to standards down to individual products.

\subsection{Industry practice for equipment selection in process design}

Design and construction of an asset like a process plant involves contributions from many specialists. System design initiates the design process, and it often interacts with area design, sourcing and construction preparation. All engineers utilize specialized applications, and they must all ensure that their part of the total scope is compliant with regulatory requirements, design codes and industry standards. 

Companies use applications that support the execution of EPC projects and enable collaboration within the organizations. However, handover between organizations is still mainly document-based. The lack of machine-interpretable versions of industry standards prevents creation of machine-interpretable representations of assets.

Purchase requisitions, product data sheets, material data sheets, material selection rules and piping class sheets are project documents that may be converted to a machine-interpretable and executable format.

\section{Engineering design documents and international standards - The valve use case}
\label{sec:designdocs}

\begin{figure}[ht]
    \centering
    \includegraphics[width=0.9\linewidth]{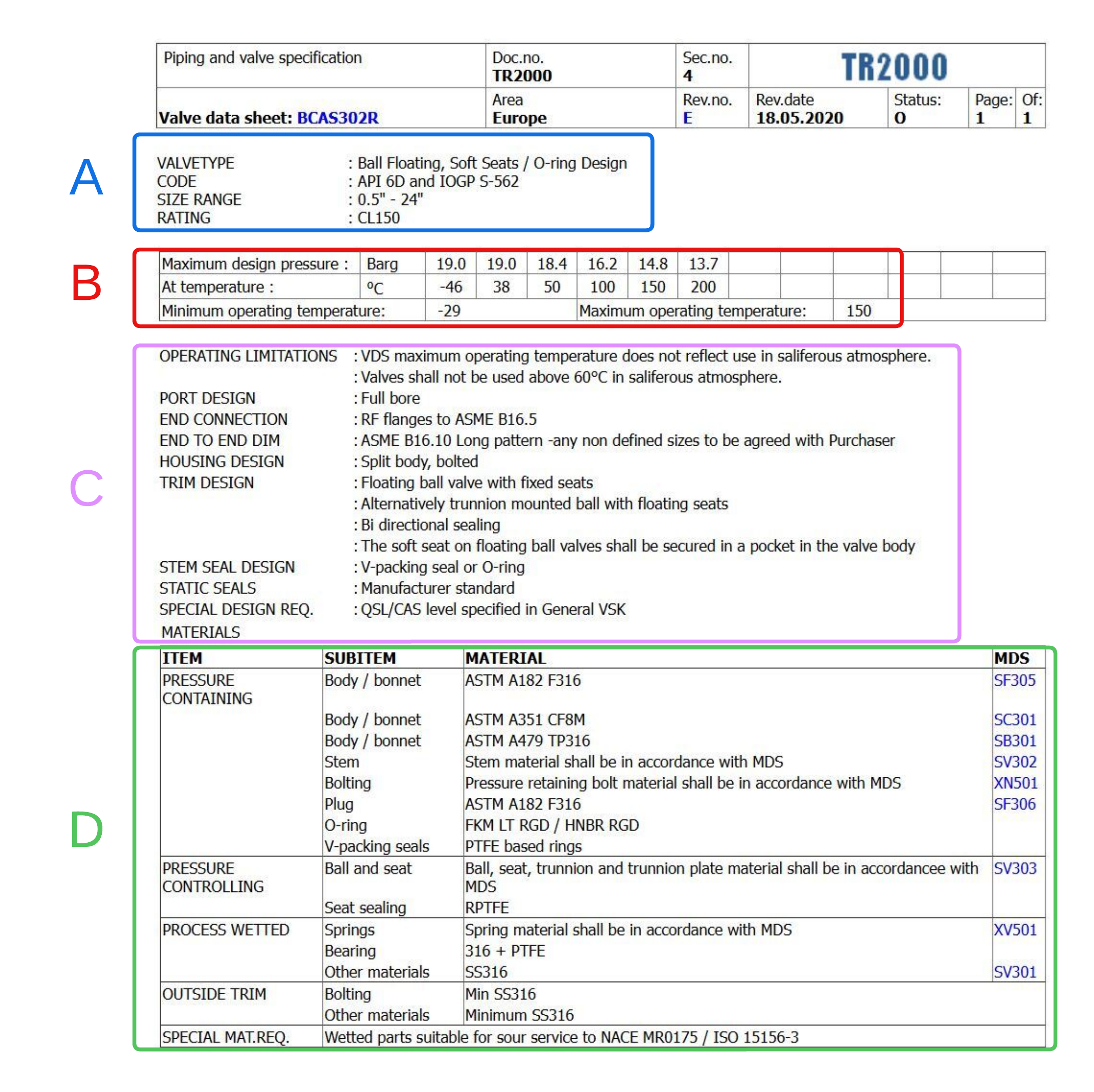}
    \caption{Example of a typical valve data sheet for full bore floating ball valve, class rating CL 150, with raised face flange connection and soft seat design showing design data and references to International API, ASME and ASTM standards}
    \label{fig:BCAS302R}
\end{figure}

In order to test that the ontology modules we create from the engineering standards are machine-interpretable and can be used to automate the Engineering Design quality assurance process we test these modules on a use case. Offshore facilities and chemical process plants are high hazard facilities and can have thousands of valves. Each of these valves has a purpose in the plant system design. Valve type selection is done for each valve, regardless of valve size and rating, to ensure that the valve is fit for purpose given the design conditions and its purpose. Ensuring correct Valve Data Sheet (VDS) is a safety critical activity and key design assumptions must be documented in the asset's Safety Case submission to the regulating authority. The VDS is a valve design specification. \cite{smith2004valve, skousen2006valve}. 

The use case considers two valve data sheets from different organisations and a valve product from a vendor. One of these VDS, for a full bore floating ball valve, class rating CL 150, with raised face flange connection and soft seat design is shown in Figure \ref{fig:BCAS302R}. It is publicly available on its web site at \url{https://tr2000.equinor.com/index.jsp}. We refer to this by its VDS number BCAS302R. We have developed ontologies and processes to automate the checks that 1) this VDS is suitable for a specific application and 2) that a particular valve make and model is compliant with the VDS, which is a key focus of this paper. To guide the reader in some of the key terms and concepts of interest, we describe each part of the VDS in more detail below. 

Section A has key information about the valve type (floating ball valve with soft seats), class rating (CL150), size range, and international standard compliance requirements. This valve must comply with API Specification 6D (Specification for Pipeline and Piping Valves) and IOGP S-562. (Refer to Table \ref{tab:standards} for details.) 
IOGP S-562 defines a minimum set of purchase requirements, additional to API 6D, and is not part of this paper.

Section B presents requirements for the pressure and temperature ranges (from ASME B16.34) for which this valve is suitable. For example, if the fluid in the line has a maximum design temperature of 50\degree C, then the maximum design pressure shall be no more than 18.4 Barg. 

Section C describes physical design and dimensional requirements such as Raised Face (RF) flange to ASME B16.5 (Pipe Flanges and Flanged Fittings) and long pattern dimensions to ASME B16.10 (Face-to-Face and End-to-End Dimensions of Valves). 

Section D describes requirements for the use of specific materials for specific parts of the valve. The materials for pressure containing valve body and bonnet parts must be compliant with ASTM A182 or ASTM A351 CF8M or ASTM A479 TP316. MDS stands for Material Data Sheet. SS316 stands for 316-stainless steel.

We can see from this example that one single valve can refer to as many as 20 standards and data sheets. This number can be even higher when references in referred documents and exceptions are considered. In the section below, we describe our approach to semantic modelling of information from tables in API, ASME, ASTM and other standards so that the ontology modules can be used, and reused, in the valve design specification process.

\section{Approach}
\label{sec:Approach}

This section describes how we select and structure the ontology modules produced from the engineering standards for the demonstration use case. Numerous works have been performed over the decades on methodologies, guidelines, and tools for ontology engineering, but despite this there are many open questions when starting a new ontology project \cite{poveda2022lot}. We have identified a set of conditions that typically apply - or should apply - when initiating and ontology project. Some of these conditions constitute success criteria, and addressing them should be given careful consideration. 

\begin{itemize}
    \item C1: There is a business need for developing this ontology pipeline for use in quality assurance in the engineering design and product selection processes. 
    \item C2: There is a strategic intention for international standards bodies to create and provide machine-interpretable standards. 
    \item C3: There are multiple stakeholders both within and across organisations involved in the design process so the ontologies must be shareable and interoperable using Linked Data. 
    \item C4: There must be infrastructure to create, run, maintain, store and quality control updates to the ontologies.
    \item C5: There is limited need for semantic concept development of individual classes as the ontologies are created based on information where the meaning of most terms are well defined in the industry standards. 
    \item C6: The volume of class and instance data we have to capture is large, for an ontology, so we must have the infrastructure to perform reasoning in a sensible time frame. 
    \item C7: The value of ontologies increases with their uptake in the industry so modularisation and reusability are important.
    \item C8: Domain ontologies are required to align terms across industry standard ontologies, such domain ontologies must be shared publicly.
    \item C9: Industry standard ontologies must be published in such a way that both the users need for data and the intellectual rights of the standards body are fulfilled.
\end{itemize}

In order to demonstrate the validity of the ontology modules produced from engineering standards we use these modules in a complex real-world commercial use case for valve selection. This is described in Section \ref{sec:ValveOntUseCase}. 

\subsection{Hierarchy of ontology modules}

\begin{figure}[ht]
    \centering
    \includegraphics[width=0.6\linewidth]{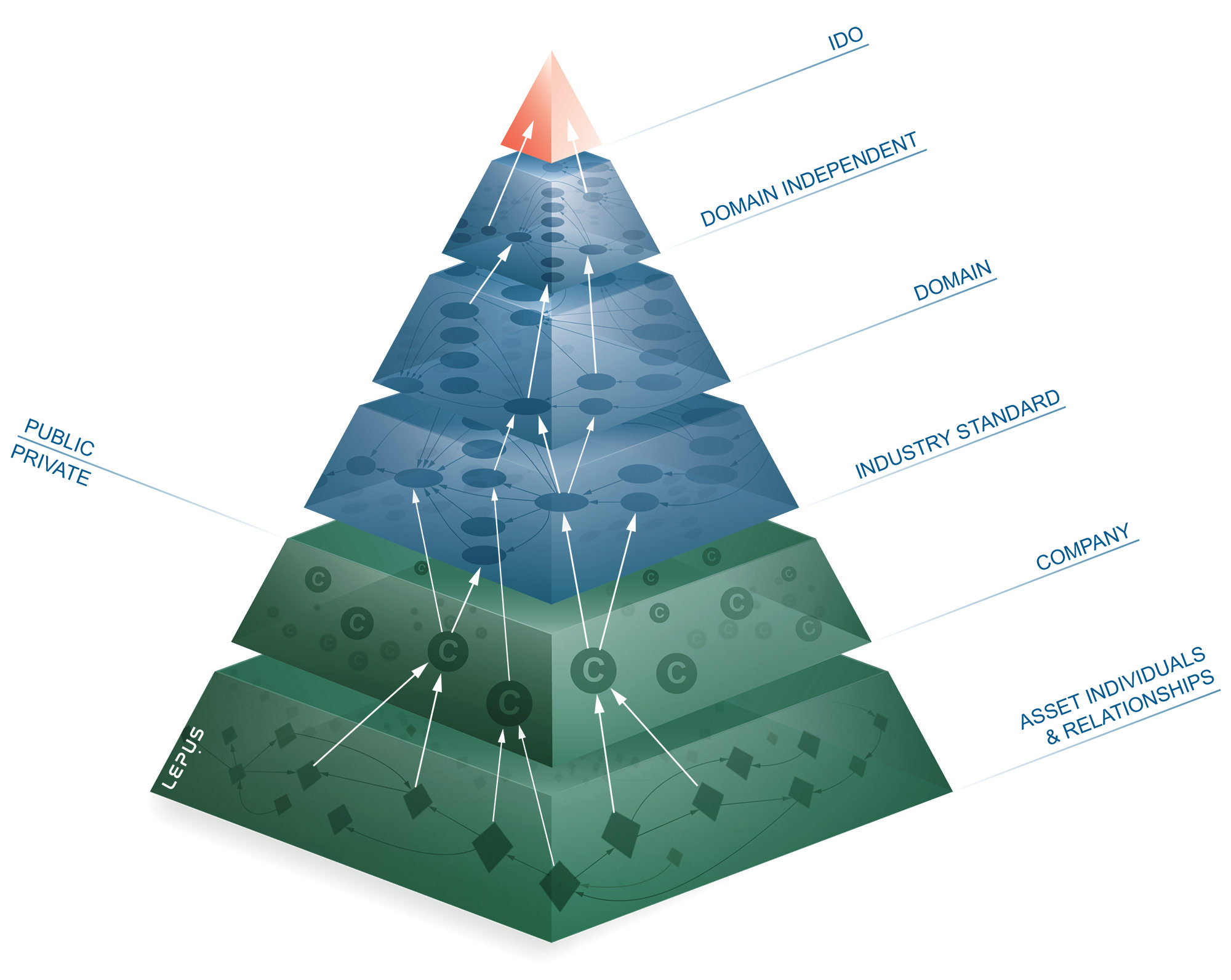}
    \caption{Industrial ontology design pyramid}
    \label{fig:pyramid}
\end{figure}

The modular ontology structure we adopt follows the import structure conceptualised in Figure \ref{fig:pyramid}. The top-level, ISO DIS 23726-3 Industrial Data Ontology (IDO) is used. IDO is part 3 of a new set of Ontology-based interoperability standard series (ISO 23726) and will be published as a full ISO standard in 2026. IDO has been used by process plant modellers, especially in the oil and gas sector, for the last decade as ISO TR 15926-14. A top level ontology, by design, contains a limited number of entities and properties. IDO contains 49 classes, 92 object properties and 2 data properties. It is designed to be used by ontologists, and needs to be extended using domain independent, domain and application ontologies to address real world questions. The modelling in these additional ontologies must adhere to the ontological commitments in the top level ontology for reasoning to work. This is achieved with the assistance of predefined, reusable, modelling patterns. Use of an agreed top level ontology is a crucial decision in modular ontology design and when semantic interoperability between stakeholders is required. 

Modelling information about equipment in process plants requires use of many common concepts and terms. As these are domain independent there are already existing, well maintained, open access modules (or terms in modules) that can be re-used providing ontological commitments are observed. Ontologies that are often imported include \textit{skos} \footnote{\url{http://www.w3.org/2004/02/skos/core}}, \textit{foaf} \footnote{\url{http://xmlns.com/foaf/0.1/}} and \textit{dcterms} \footnote{\url{http://purl.org/dc/terms/}}. 
Technical ontologies relevant to the engineering field include, but are not limited to, the \textit{Semantic Sensor Network (SSN)} \footnote{\url{https://www.w3.org/TR/vocab-ssn/}} and \textit{GeoSPARQL} \footnote{\url{https://docs.ogc.org/is/22-047r1/22-047r1.html}} ontologies.    

The next layer in the ontology pyramid is for domain ontologies. Domain ontologies contain textbook type terms, dictionaries and information types specific to a domain. For example, there may be domain ontologies for terms relating to piping and describing types of valves. Once made, these modules and can be reused by others in the domain as long as the concepts are modelled consistently with the top level ontology.  

Ontologies based on the contents of engineering standards are created as specializations of domain ontologies. The text and tables in standards documents are made for human consumption and the information is not explicit enough for direct use in ontologies. The format of tabular input to ontology generation will therefore not have an exact match to those found in the standard document. Our ambition is to capture the intent of the standard such that its ontology version can be utilized by both man and machine. Using the machine-interpretable standard ontology should give the same result as would result from a qualified engineer using the standard. 

Industry standards are version controlled. The ontology representation of standards also needs to be version controlled. Resources in the industry standard ontology, typically OWL classes, will have references to one or more versions of the standard document. However, such references should be included as annotation properties only, to avoid undesirable inheritance. For unchanged parts of a standard the ontology representation can thereby remain unchanged across multiple versions of the standard document.

Experience based preferences, local conditions, regulatory requirement and equipment availability are among the reasons why companies create their own specifications. Such specifications are to a large extent specializations of industry standards and can be converted to ontologies. The organization owning the specification will also control any sharing of any ontology version of the specification. The ontology pyramid shown in Figure \ref{fig:pyramid} allows for  collaboration about specifications in a controlled manner.

Asset individuals (instance data for the specific asset) are at the bottom of the Figure \ref{fig:pyramid}. Asset information often need to be kept private, but when collaboration is required parts of asset data can be shared along with relevant semantics.

\subsection{Modelling patterns}

When software and ontology engineers model semantic data in a domain they make numerous decisions including what concept is being captured, the label it uses, how one concept is related to another, and what label to give this relation. Different perspectives and abstractions inform choices and without control the result is a proliferation of idiosyncratic logical models that cannot easily be used or reused. Hence, modelling patterns are as important as the ontologies themselves to enable interoperability. Using the same modelling patterns across industry standards contributes to the alignment of industrial asset information. For example, to enable comparison between specified requirements (quantified in terms of values and units of measure) with asset or equipment properties, these properties must be formatted such that they may be compared. Semantic specification of objects will typically include multiple object property relationship types in addition to classification. The use of published and documented common machine-executable templates is a necessary part of an industrial semantic ecosystem \cite{skjaeveland2018semantic}.

Ontology modelling is inherently complex and the use of semantic web languages for modelling industrial data has not been common practice amongst software developers. Ontology modelling patterns present a way forward to aid developers by providing documented, quality controlled, off-the-shelf building blocks for software engineers to use\cite{blomqvist2016considerations}. 

In this work we use the OTTR framework for pattern-based ontology engineering. OTTR is a generic templating language for representing and instantiating modelling patterns by nested and parameterized templates. It is described in a number of papers, most recently and completely in \cite{skjaeveland2024reasonable} and is open access \footnote{https://ottr.xyz/}. OTTR templates are documented and can be shared as template libraries for different users at different levels of abstraction and then instantiated using the OTTR tool kit.
OTTR is also a community supported by a shared library of reusable patterns, tools for practical pattern instantiation and maintenance, and active users. As OTTR supports the construction and use of large-scale ontologies and knowledge graphs serialised in RDF it is being used by a number of industrial organisations including Aibel, a global engineering company \cite{skjaeveland2018semantic}, Grundfos, a global pump manufacturer \cite{brynildsen2023building}, Bosch, a global product manufacturer \cite{svetashova2020ontology}, and the Norwegian Maritime Authority \cite{heimsbakk2023using}.  All of the ontologies used in this project are built from ontology (re-usable) OTTR patterns created initially by engineers using an tabular (Excel) interface. 

\begin{figure}[h!]
    \centering
    \includegraphics[width=1.0\linewidth]{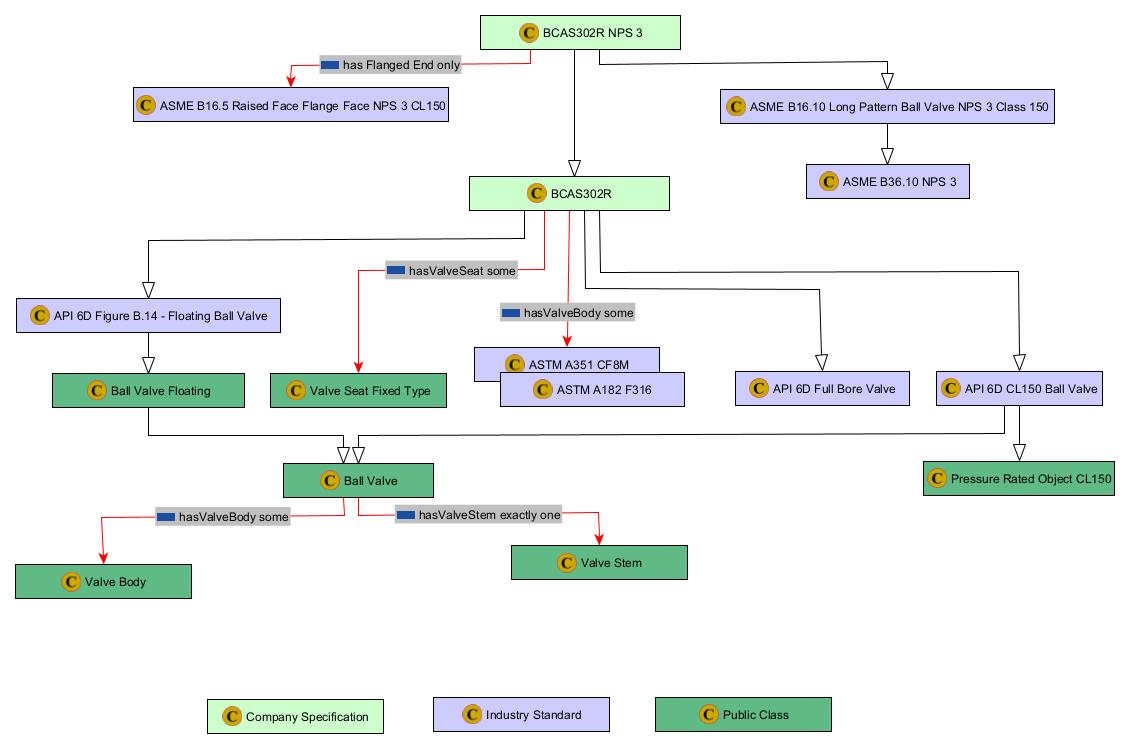}
    \caption{Classes and object properties for the BCAS302R Floating Full Bore Valve whose datasheet is shown in Figure \ref{fig:BCAS302R}. Colours are used for different ontology classes as follows: green for valve parts and light green for the use case. Black lines with open arrows represent subclass relations and red lines are object properties with labels in grey boxes.}
    \label{fig:valve model}
\end{figure}

A visual example of a pattern for modelling the physical structure of one valve (a `Floating Full Bore Ball Valve with Fixed Seats or Trunnion Mounted with Floating Seats') is shown in Figure \ref{fig:valve model}. Note that only the option with floating ball is shown. This module (and the associated code, not shown) can be used and reused to ensure consistency in how such a model is conceptualised. 

\subsection{Semantic reasoning and design rule compliance}

Quality assurance in engineering design is fundamentally a process of if-then rule checking. For example, if the \textbf{Stream} \textbf{containedBy} the \textbf{Valve On-off} is \textbf{Potable Water} AND nominal size is \textbf{DN 80 - NPS 3} AND it \textbf{hasSpecifiedMaxOperatingPressureBarg} < 10 AND \textbf{hasSpecifiedMaxOperatingTemperatureDegC} < 50 then there is a constrained search of a decision tree to identify a potential set of valves that might be suitable.

\begin{itemize}
\item \textbf{Stream}                                  http://rds.posccaesar.org/ontology/lis14/rdl/Stream
\item \textbf{containedBy}                             http://rds.posccaesar.org/ontology/lis14/rdl/containedBy
\item \textbf{Valve On-off}                            http://data.aibel.com/rdl/X101609012
\item \textbf{Potable Water}                           http://data.aibel.com/rdl/X101633249
\item \textbf{DN 80 - NPS 3}                           http://data.aibel.com/rdl/X101004239
\item \textbf{hasSpecifiedMaxOperatingPressureBarg}    http://data.aibel.com/rdl/X101626125
\item \textbf{hasSpecifiedMaxOperatingTemperatureDegC} http://data.aibel.com/rdl/X101626105
\end{itemize}

Two engineers given the same set of conditions and the same set of standards should arrive at the same answer. If we can do this with people, we can do this with a machine, using ontologies and their reasoning ability. 

Deciding what is in and out of individual modules depends in part on the type of ontology. An important decision is how to capture rules that go beyond the subclass relationship and constraints associated with domain and range on object properties. For example, should logical axioms such as requiring that a ball valve must have one ball closure member as shown in the code below be in a domain ontology such as \textsf{valve-core} or in a company or use case ontology? Our view is that logical axioms should be kept in a built-for-purpose ontology module but where they are present in tables in industry standards, should be included in the ontology modules developed for those standards.

\begin{verbatim}
    <http://data.aibel.com/rdl/X101037403>
    rdf:type owl:Class ;
    rdfs:subClassOf <http://data.aibel.com/rdl/X101037400> ,
    <http://data.aibel.com/rdl/X101609104> ;
    idov:inOntology "valve-core" ;
    rdfs:comment "A rotary valve that has a ball closure member" ;
    rdfs:label "Ball Valve" ;
    owl:qualifiedCardinality "1"^^xsd:nonNegativeInteger.
\end{verbatim}

For the application ontology (use case), stages at which design rules are executed should, as a minimum, match stages at which manual checking activities in document based work processes are executed. Engineering rules, as shown in Section C of Figure \ref{fig:BCAS302R}, for example requiring that material for the body and bonnet of the pressure containing part of the BCAS302R ball valve must be of material ASTM A182m A351 or A479 are only captured in the BCAS302R application use case ontology module.   

\subsection{Ontology testing}
The valve-core ontology was tested along with its import of the IDO core ontology using OOPS Ontology Pitfall Scanner \cite{poveda2014oops}. Only minor issues were identified. A copy of the scanner results is available in the GitHub \footnote{https://github.com/PCA-POSC-Caesar-Association/IDO\_Use\_Case\_Valve\_Ontology\_Public}.

\section{Creating ontology modules from engineering standards}
\label{sec:valve_ont_modules}

This section describes decisions, processes and tools we use or develop to create the components and structures used in valve ontology use case described in section \ref{sec:ValveOntUseCase}. We concentrate in particular on the creation of the valve core module and the ontology modules from specific industry standards required in valve specification (namely ASME B16.34 and API 6D) showing examples of how information contained in these documents can be modelled as individual ontology models.

\subsection{Valve core module}

\begin{figure}[h!]
    \centering
    \includegraphics[width=1.0\linewidth]{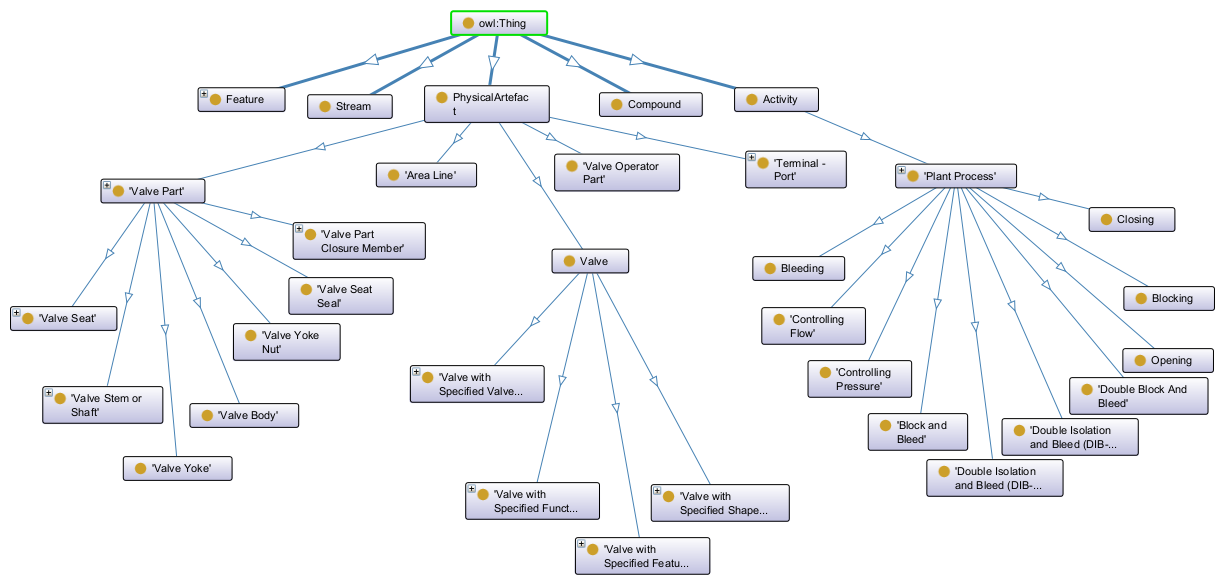}
    \caption{Top-level classes in the Valve Ontology file, from the Protégé Editor. }
    \label{fig:valve core}
\end{figure}

The aim of the valve core ontology is to create classes and object properties for terms used to describe valve-related terms and hierarchical structures commonly used by the engineering community. Valve related terms include, for example, valve types and parts. 
There are many different types of valves and selection of an appropriate valve includes consideration of function and form (design) \cite{smith2004valve}. The ontology does not constrain the user to a single hierarchy as often happens in table of contents in textbooks, catalogues and standards \cite{nesbitt2011handbook,skousen2006valve}. We model a valve as an equivalent \textit{Valve} class with subClasses as follows: \textit{Valve with Specified Function} (e.g. Fail Close Valve, Three Way Valve), \textit{Valve with Specified Feature} (e.g. Valve with Full Bore), \textit{Valve with Specified Shape} (e.g. Valve Three-way, Valve Long Pattern)  and \textit{Valve with Specified Valve Type} (e.g. Ball, Butterfly, and Gate Valves, each may also have subclasses).  These are shown in Figure \ref{fig:valve core}. 

\begin{figure}[h!]
    \centering
    \includegraphics[width=1.0\linewidth]{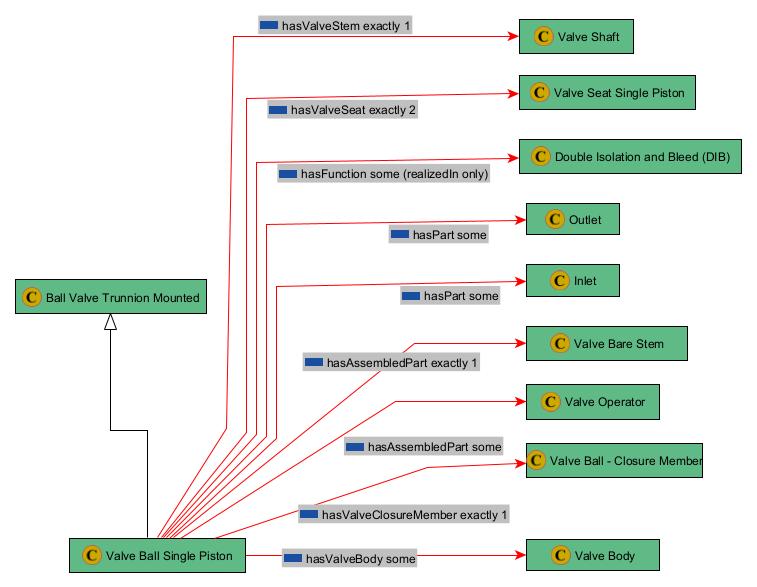}
    \caption{Object properties in the Valve Core Ontology}
    \label{fig:valve object properties}
\end{figure}

Valves have many of the same generic parts and these are assigned subclasses to \texttt{Valve Part} such as a \texttt{Valve Stem}, \texttt{Valve Body}, \texttt{Valve Bonnet} and \texttt{Valve Cover}. These classes are used to create re-usable modelling patterns for specific valve types. For example a \texttt{Ball Valve} can be modelled by subproperties of \textit{hasAssembledPart} (an object property in IDO) such as \textit{hasValveStem, hasValveBody} creating triples such as \texttt{Ball Valve}-\textit{hasValveStem}-\texttt{Valve Shaft}. A diagram illustrating these specialised object properties is shown in Figure \ref{fig:valve object properties}. Using these object properties is a more efficient way of modelling than giving each valve design separate classes for each part such as \texttt{ball valve stem}, \texttt{gate valve stem} and so on, as is done in some industry reference data libraries. Features that are specific to a particular valve design such as \texttt{Ball Valve Floating} and \texttt{Ball Valve Trunnion Mounted} are given their own classes with the valve type (e.g. ball) as a prefix. 

In addition to the creation of sub-properties of \textit{hasAssembledPart} mentioned above, these sub-properties are given domain and range restrictions. For example, \textit{hasValveBody} with domain \texttt{Valve} and range \texttt{Valve Body}. The approach captures what engineers know to be true, namely that a valve only has one body  (e.g. \textit{hasValveBody} exactly 1 \texttt{Valve Body}) and enables specification of valve body properties within the VDS class. The same approach is used for other typical parts and features. 

The intention is that this valve core module contains the valve-related terms necessary for modelling the contents of the valve-related engineering standards, specifically API 6D and ASME B16.34, as well as the information used in the valve design specification process. By doing this, the Linked Data and annotations created can be reused, making for a more efficient process and reducing the potential for duplication of classes or creation of alternative and incompatible hierarchies. All the valve-related terms developed for the valve ontology are stored in the \textsf{valve-core} module file. Valve core contains 197 classes, 30 object properties, and 17 individuals. The \textsf{valve-core} module models a valve, its components, and other physical elements to which it is connected. 

A key feature of valve engineering standards are the tables containing values with associated units of measure associated with, for example, dimensions (minimum bore size mm) and pressure (minimum operating pressure barg). To assure effective semantic reasoning OWL Data properties are used for numeric values. The data properties have annotation type references to an industry reference data library of Open Linked Data called the PLM RDL (available at \url{https://rds.posccaesar.org/ontology/plm/}). An example of an Open Linked Data address for pressure is \url{https://rds.posccaesar.org/ontology/plm/rdl/PCA\_100003596/}

The \textsf{valve-core} module has only a single import, the IDO ontology. It is not dependent on any of the other imports shown in Figure \ref{fig:valveont-imp}. Valve-core is built from tabular input using templates.

\subsection{Modules for the ASME B16.34 Valve standard}

The ASME standard B16.34 Valves — Flanged, Threaded, and Welding End covers technical requirements for valves specifically pressure–temperature ratings, dimensions, tolerances, materials, nondestructive examination requirements, testing, and marking for cast, forged, and fabricated flanged, threaded, and welding end and wafer or flangeless
valves of steel, nickel-base alloys, and other alloy. Pressure–temperature ratings for valves are an important part of this standard. A key modelling challenge in ASME B16:34 is how to model the working pressures by temperature for each standard Piping Class and capture the acceptable materials standards (described by a Group classification). We have created the ASME B16.34 PT ontology to represent the Pressure-Temperature data so that this data becomes suitable for design validation purposes using semantic reasoning.

ASME B16.34 PT has 49 material groups each with a list of ASTM material grades allowable for use in pressure-retaining structural parts of the valve. 
For each material group allowable maximum pressure at a maximum temperature is stated for pressure class CL150, CL300, CL600, CL900, CL1500, CL2500 and CL4500. Additional to this the standard specifies A – Standard Class, B – Special Class within each Material Group with specific pressure temperature ranges for each pressure class. In the B16.34 ontology, Material Group 2.2 is selected as example of a Material Group, as shown in Figure \ref{fig:ASMEB16.34_2.2}. The class ASME B16.34 Material Group 2.2 Object is stated to be equivalent to:

\small
\begin{lstlisting}
 ('ASTM A 312 Grade TP316 Compliant Object' or 'ASTM A 358 Grade 316 Compliant Object' or  'ASTM A 182 Grade F316 Compliant Object' or  'ASTM A 240 Grade 316 Compliant Object' or 'ASTM A 351 Grade CF8M Compliant Object' or 'ASTM A 182 Grade F316H Compliant Object' or 'ASTM A 182 Grade F317 Compliant Object' or 'ASTM A 240 Grade 316H Compliant Object' or 'ASTM A 240 Grade 317 Compliant Object' or 'ASTM A 312 Grade TP316H Compliant Object' or 'ASTM A 312 Grade TP317 Compliant Object' or 'ASTM A 351 Grade CF3A Compliant Object' or 'ASTM A 351 Grade CF8A Compliant Object' or 'ASTM A 351 Grade CG8M Compliant Object' or 'ASTM A 376 Grade TP316 Compliant Object' or 'ASTM A 376 Grade TP316H Compliant Object' or 'ASTM A 351 Grade CF10M Compliant Object' or 'ASTM A 351 Grade CG3M Compliant Object' or 'ASTM A 430 Gr. FP316 Compliant Object' or 'ASTM A 430 Gr. FP316H Compliant Object' or 'ASTM A 479 Grade 316H Compliant Object')
 \end{lstlisting}

\normalsize

All the ASTM-compliant Objects are ASME material grades represented as OWL classes. \texttt{ASME B16.34 Material Group 2.2 Object} is a classification for valve component material with no reference to pressure or temperature.
The \texttt{ASME B16.34 Material Group 2.2 Valve} class includes valves with \texttt{ASME B16.34 Material Group 2.2 Object} in pressure retaining parts. The object property relationships \textit{hasValveBody} only \texttt{ASME B16.34 Material Group 2.2 Object} and \textit{hasValveBonnetOrCover} only \texttt{ASME B16.34 Material Group 2.2 Object} are class restrictions on the \texttt{ASME B16.34 Material Group 2.2 Valve}. 

The content of the pressure temperature table for Material Group 2.2 is captured using OWL classes. There is one class representing each allowable temperature range and one representing each allowable pressure range. Each OWL class is restricted using data property types, namely \textit{hasSpecifiedMaxDesignTemperatureDegC} and \textit{hasSpecifiedMaxDesignPressureBarg}. The OWL class \texttt{WP <= 1 barG} is a sub class of \texttt{Object with working pressure} and \texttt{WT <= 816 deg.C} is a sub class of \texttt{Object with working temperature}. Each conjunction set of classes are included, e.g. \texttt{WP <= 18.4 barG} and \texttt{WT <= 50 deg.C} or  \texttt{WP <= 19 barG} and \texttt{WT <= 38 deg.C}.

\texttt{ASME B16.34 Material Group 2.2 Valve CL150 A– Standard Class} is a sub class of \texttt{ASME B16.34 Material Group 2.2 Valve} and \texttt{ASME B16.34 CL 150 Pressure Rated Object}. It is also defined as equivalent to:

\small
\begin{lstlisting}
 ('WP <= 1 barG' and 'WT <= 816 deg.C') or ('WP <= 1.2 barG' and 'WT <= 800 deg.C') or ('WP <= 1.4 barG' and 'WT <= 538 deg.C') or ('WP <= 1.4 barG' and 'WT <= 550 deg.C') or ('WP <= 1.4 barG' and 'WT <= 575 deg.C') or ('WP <= 1.4 barG' and 'WT <= 600 deg.C') or ('WP <= 1.4 barG' and 'WT <= 625 deg.C') or ('WP <= 1.4 barG' and 'WT <= 650 deg.C') or ('WP <= 1.4 barG' and 'WT <= 675 deg.C') or ('WP <= 1.4 barG' and 'WT <= 700 deg.C') or ('WP <= 1.4 barG' and 'WT <= 725 deg.C') or ('WP <= 1.4 barG' and 'WT <= 750 deg.C') or ('WP <= 1.4 barG' and 'WT <= 775 deg.C') or ('WP <= 2.8 barG' and 'WT <= 500 deg.C') or ('WP <= 3.7 barG' and 'WT <= 475 deg.C') or ('WP <= 4.6 barG' and 'WT <= 450 deg.C') or ('WP <= 5.5 barG' and 'WT <= 425 deg.C') or ('WP <= 6.5 barG' and 'WT <= 400 deg.C') or ('WP <= 7.4 barG' and 'WT <= 375 deg.C') or ('WP <= 8.4 barG' and 'WT <= 350 deg.C') or ('WP <= 9.3 barG' and 'WT <= 325 deg.C') or ('WP <= 10.2 barG' and 'WT <= 300 deg.C') or ('WP <= 12.1 barG' and 'WT <= 250 deg.C') or ('WP <= 13.7 barG' and 'WT <= 200 deg.C') or ('WP <= 14.8 barG' and 'WT <= 150 deg.C') or ('WP <= 16.2 barG' and 'WT <= 100 deg.C') or ('WP <= 18.4 barG' and 'WT <= 50 deg.C') or ('WP <= 19 barG' and 'WT <= 38 deg.C')
 \end{lstlisting}
 \normalsize
 
In the valve specification process the ASME B16.34 ontology is used to verify material selection. For a specific valve at a location in a plant the ontology can be used to determine suitability of a valve product given the design pressure and temperature at the location. To achieve this we create a number of  ontology modules. Each of these modules is available in the GitHub repository that accompanies this paper at \url{https://github.com/PCA-POSC-Caesar-Association/IDO_Use_Case_Valve_Ontology_Public}. 

\subsection{API 6D standard and 602 standard}

The API 6D international standard specifies requirements and provides recommendations for the design, manufacturing, testing and documentation of ball, check, gate and plug valves for application in pipeline systems  for the petroleum and natural gas industries. The standard establishes terminology for describing these valves, their parts, configurations, dimensions, features, materials and testing requirements. We have concentrated for this work on how to represent key information pertaining to Class 150, 300, 600, 900, 1500 and 2500 valves (ball, check, gate and plug) across the range of NPS sizes and for different mounting options, design and body configurations.

The API standard 602 covers three specific valve types (Gate, Globe and Check valves) and is only for sizes NPS4 and smaller for the petroleum and natural gas industries. The ontological representation is similar to the API 6D representation.

\subsection{The valve-collect ontology module}

As industrial ontologies are used by numerous stakeholders we have developed a simple naming criteria as follow. Various ontologies necessary for a domain are `collected' into a \textsf{collect} file. In this case we have \textsf{valve-collect}, \textsf{piping-collect} and \textsf{materials-collect} files. These ontology modules contain no classes of their own but import existing modules. The \textsf{valve-collect} imports five modules based on the engineering standards API 6D, API 602 and ASME 16.10. 

Next we have a set of `core' files such as \textsf{valve-core, piping-core} and \textsf{materials-core}. These contain defined and commonly used concepts, taxonomies and relationships used by engineers and in engineering standards when describing piping, materials and valves.  
Where necessary `core' files import other `core' files, so \textsf{valve-core} imports \textsf{material-core} and \textsf{ piping-core}. Both of these are extracts from Aibel \textsf{MMD-core}, a proprietary ontology module. We have already covered valve core and describe the other modules briefly below.

\subsection{Piping and materials ontology modules}

Terms and relationships that are specific to a domain and are used in engineering standards are stored in `core' files namely \textsf{piping-core}, \textsf{materials-core}, and \textsf{MMD-core}. \textsf{MMD-core} captures material master data and imports two modules \textsf{control} and \textsf{annulus} for discipline-specific and shape/dimension terms respectively. All of these are commercially-sensitive so cannot be provided as open data but we describe features of them briefly in the subsections below.


\subsubsection{Piping ontology modules}

The ontology module \textsf{piping-collect} is an ontology based on collecting ontology modules of piping domain features. It has no classes of its own. The \textsf{ASME\_B16\_5\_FEA} is included to enable classification of flanged end connections of valves valve. ASME B16.5 includes a selection of flange features with size, pressure rating and end types (Ring Type Joint, Raised Face or Flat Flange Face) The use case version of the ontology has 72 classes and 260 axioms. An industrial ASME B16.5 feature version has more then 450 classes and 5000 axioms and imports the ontology module \textsf{piping-core}.

The ontology module \textsf{piping-core} captures generic content related to piping. This includes generic product types and feature types that is used in multiple standards. The version of \textsf{piping-core} used in the use case includes end features for valves, and no generic product types. The use case version of the ontology has 131 classes and 1097 axioms. An industrial piping-core version has more then 1400 classes and 10000 axioms. There is no dimensional or materials data in \textsf{piping-core}.

\subsubsection{Materials ontology modules}

The \textsf{materials-collect} ontology combines all the materials domain ontologies importing ontology modules for ASTM A182, A240, A312, A351, A358, and A376 as well as the \textsf{materials-core} file. Like the \textsf{piping-collect} ontology it has no classes of its own.

The \textsf{materials-core} module contains classes organised by ASTM standards numbers (e.g. ASTM A376) and a taxonomy of materials (e.g.carbon steel, elastomer) as well as quality factors artefacts from standards such as ASME B31.3 and ASTM A928. Like the \textsf{piping-core} file, this module, and the associated ASTM ontology modules were not created specifically for this project but were existing amongst the project partners and have been reused.

\subsection{Provenance}

An important consideration for engineers is to check that the `right' version of an engineering standard is being used. To achieve this we make use of an annotation property with the label \textit{isDefinedInEditionOfSpecification}. This identifies the relationship from class or individual to a version of a specific standard. One of the examples (see the EquinorMyPlant.owl file in the GitHub \footnote{\url{https://github.com/PCA-POSC-Caesar-Association/IDO_Use_Case_Valve_Ontology_Public}} is the `ASME B16.10 Long Pattern Ball Valve NPS 10 Class 150 RF \texttt{isDefinedInEditionOfSpecification} `ASME B6.10 Face-to-Face and End-to-End Dimensions of Valves ASME B16.10-2009'.

\section{Valve ontology use case instantiation}
\label{sec:ValveOntUseCase}

The value of the ontology modularisation approach described in Section \ref{sec:Approach} and the creation of individual modules from valve-related engineering standards in \ref{sec:valve_ont_modules} is assessed through their use in a valve selection use case.  We provide a copy of the valve core ontology and examples to support the use case described below at \url{https://github.com/PCA-POSC-Caesar-Association/IDO_Use_Case_Valve_Ontology_Public}. As mentioned in Section \ref{sec:designdocs} the use case considers two valve design specifications from different organisations and a valve product from a vendor.

Ontology evaluation aims to ensure that the ontologies represent the intended domain knowledge, are logical consistent, and support solvable, reliable and meaningful inference. Ontology evaluation is usually done with respect to a frame of reference, such as competency questions based on use cases. We consider the following use case with two VDS, one is the BCAS302R valve from Equinor's TR2000 VDS series and the other is the valve VDS AB-GTDD00J from Aker BP.  Some details of each Use Case are in Table \ref{tab:ValveInfo} and we will refer to these instances by their identifiers in the remainder of the paper noting that the VDS are specifications (not valves), the proposed location ID is associated with a functional specification for pressure, flow etc, and only the proposed product specification represents an actual valve that could be purchased. 

We demonstrate reasoning using the valve ontology applied to real industry data. There are two competency questions:
\begin{enumerate}
    \item Is this VDS suitable for the specific functional application? We have two applications and two VDS.
    \item Is the proposed valve product's make and model compliant with the VDS for each use case?
\end{enumerate}

\begin{table}[ht]
\begin{tabular}{p{6cm}p{5cm}p{4cm}}
                     & Equinor Use Case      & Aker BP Use Case    \\ \hline
Location ID          & P-63-CW032            & A-64GT0073          \\
Pipe Class           & AS200                 & DD20                \\
VDS ID               & BCAS302R              & AB-GTDD00J          \\
Source               & Equinor VDS           & Aker BP VDS         \\
Valve type           & Ball Floating         & Wedge gate          \\
Size                 & DN 80-NPS 3           & DN 20-NPS 3/4       \\
Pressure Rating      & CL150                 & CL600               \\
Flange face          & RF                    & RTJ                 \\
Service              & Plant Air             & Nitrogen            \\
Purpose              & On/Off                & Drainage (PDS VD01) on purge line  \\
Line Number          & 3"-AI-63-006-AS200    & A-64L00154A-0200GI-DD20-000000N     \\
Proposed product Type     & O.M.S.SALERI S7100.SF & IKM Flux Fig.No L6RR104        \\ 
Alternate product & DAFRAM S.p.a.F1FS NPS3 CL150 & \\ \hline
\end{tabular}
\caption{Examples of process design information necessary in the development of VDS for the Equinor and Aker BP Use Cases and the names of the proposed vendor products.}
\label{tab:ValveInfo}
\end{table}

\subsection{Ontology modules and import hierarchy}

\begin{figure}[h!]
    \centering
    \includegraphics[width=1.0\linewidth]{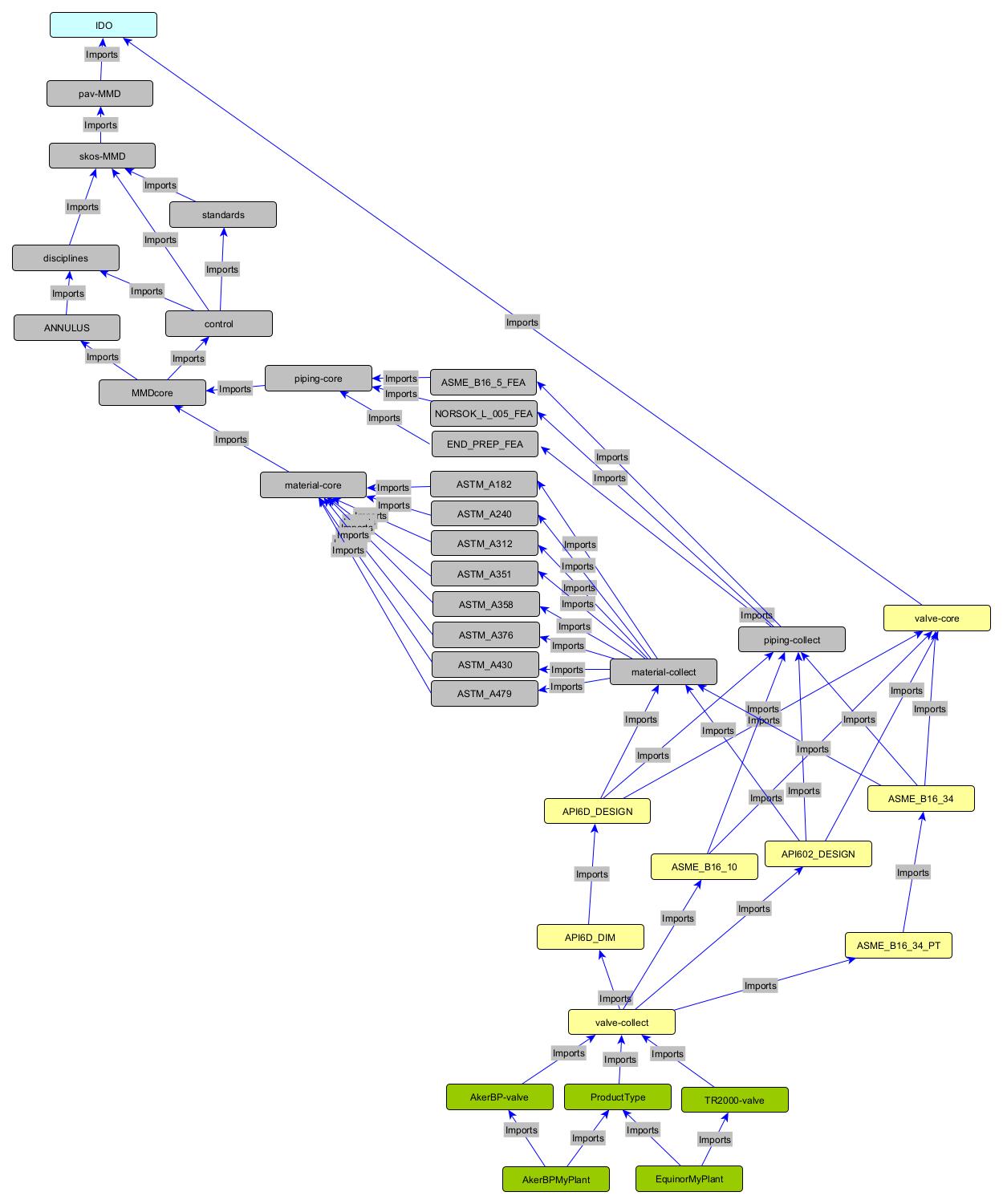}
    \caption{Ontology import structure - Those marked gray are extracts from ontologies in use at Aibel. In yellow we have ontologies created as part of the "IDO Valve use case". In green we have proposed company specific ontologies. These are created as part of "IDO Valve use case" .}
    \label{fig:valveont-imp}
\end{figure}

The ontology import structure is shown in Figure \ref{fig:valveont-imp}. This is an illustration of the principles in the Industrial Ontology design pyramid shown in Figure \ref{fig:pyramid} showing how we re-use the existing classes, object and data properties, and axioms available in industry standard, domain, domain-independent and IDO ontologies.  

As shown in Figure \ref{fig:valveont-imp} this is a complex use case involving numerous ontology modules. The diagram shows the IDO core ontology on the right hand side. Valve-core only imports IDO core. The valve-related standards we have modelled as part of this work (ASME 16.34, ASME B16.10, API6D, API 602).

It is necessary for us to make use of a number of pre-existing ontology modules from piping and materials standards in addition to the new one (valve core) we have made specifically from the API/ASME/ASTM standards for valves. These piping and materials ontology models were developed over the past decade by Aibel, an engineering design house, and have been used for quality assurance in piping design for offshore plants\cite{kluwer2008iso, skjaeveland2018semantic}.  These modules are commercially-sensitive and cannot be released. For the purposes of industry application we only need to create individuals specific for the use case. 

\subsection{Modelling the functional specification}

The first step is to build an asset model as an ontology. The asset model contains instances of plant process streams, equipment and plant areas and the topology. Data from engineering registers was used as tabular input when building both the EquinorMyPlant and the AkerBPMyPlant ontologies. The resulting asset model includes information about equipment and piping that is typically found on a Piping and Instrument Diagram (P\&ID) and it includes data properties of process lines typically found in a Line List. The plant object is classified with classes from the ontology used as reference data. Valves are fully specified to Valve Data Sheet (VDS) level utilizing the purpose-made Equinor and Aker BP specific valve ontologies, TR2000-valve and AkerBP-valve. Finally, the relationship between an instance of the valve manufacturers valve type and the valve tag is created.

\begin{figure}[h!]
    \centering
    \subfloat[\centering Ontology Model]{{\includegraphics[width=8cm]{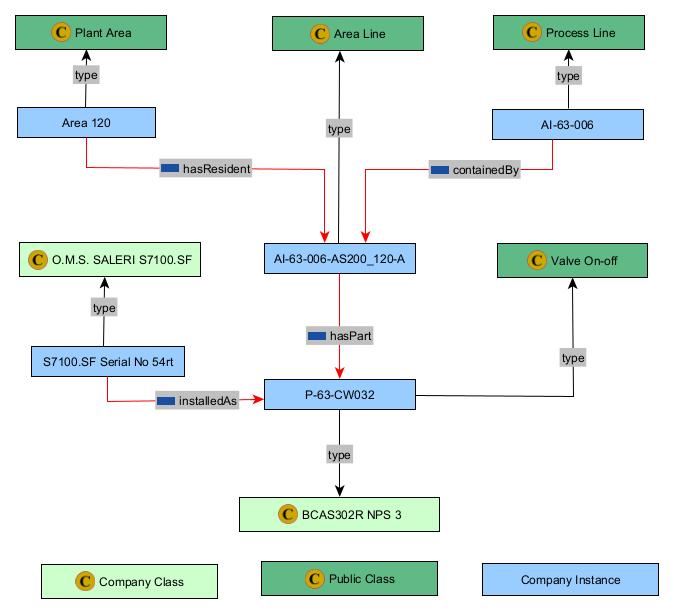}}}
    \qquad
    \subfloat[\centering Extract from P\&ID]{{\includegraphics[width=6cm]{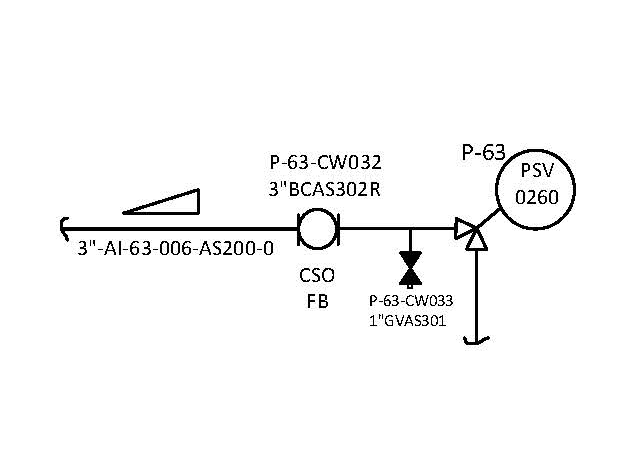}}}
    \caption{Ontology model for the functional specification of the Ball Valve P-63-CW032.}
    \label{fig:funcSpec}
\end{figure}

Figure \ref{fig:funcSpec} shows an example of the ontology model for the “BCAS302R DN 80 - NPS 3 EQUINOR VALVE” in line 3"-AI-63-006-AS200. Our intention in the use case is to check if P-63-CW032 can be implemented by the product “O.M.S. SALERI S7100.SF Serial No 54rt” valve. Valve tag P-63-CW032 is classified as “Valve on-off”

We model the features of the individual P-63-CW032 as follows in Table \ref{tab:InstanceODP}. This table illustrates both bject property and data property assertions. Making these explicit is important for subsequent reasoning.

\begin{table}[h!]
\begin{tabular}{p{6 cm}| p{8 cm} p{1cm}}
\hline
Object property   assertion      & Data property assertion     &                       \\ \hline
assembledPartOf AI-63-006\_120-A & hasSpecifiedActualDensityKGM3 & 10.4                \\ 
contains 3"-AI-63-006-AS200      & hasSpecifiedNormalOperatingVelocityms & 0            \\ 
                                 & hasSpecifiedMaxDesignTemperatureDegC & 60            \\ 
                                 & hasSpecifiedMinDesignTemperatureDegC & -7            \\ 
                                 & hasSpecifiedMaxDesignPressureBarg & 13              \\ 
                                 & hasSpecifiedNormalOperatingPressureDropBar/100m & 0   \\ 
                                 & hasSpecifiedMinDesignPressureBarg & -1                \\ 
                                 & hasSpecifiedActualMassFlowrateKg/h & 4040             \\ \hline
\end{tabular}
\caption{Specified object and data properties for the instance of the BCAS302R valve}
\label{tab:InstanceODP}
\end{table}

\subsection{Modelling each VDS}

Each VDS needs its own semantic definition. In Figure \ref{fig:valve model} we illustrate modelling of a section BCAS302R valve data sheet shown in Section D of Figure \ref{fig:BCAS302R}. Each VDS with size is modelled as an OWL class. Industry standard material grades(e.g. ASTM A 351 Grade CF3A) are also modelled as classes with restrictions. All classes in Figure \ref{fig:valve model} are already defined in ontology imports. For example the classes in the purple boxes come from ontology modules developed for the API6D (valve) and ASTM (materials) standards. The green boxes for the valve elements and the object properties such as \textit{hasValveStem} shown as red lines come from the  \textit{valve-core} ontology. The use of object property restrictions enables specification of both the valve assembly and its individual parts with on OWL class.

\subsection{Reasoning for the use cases}

A semantic reasoner infers logical consequences from the asserted axioms in an ontology. If these axioms mimic the rules used by an engineer for a specific process then the ontology can be used as an aid by the engineer. The Valve Use Case ontology has been checked for inconsistencies using the HermiT reasoner and also by an experienced Piping Engineer. We have performed a number of checks with the Protege Ontology Editor reasoner as shown in Table \ref{tab:reasonerresults}.

\begin{table}[h!]
\begin{tabular}{|p{1.5cm}|p{8 cm}|p{5.5cm}|}
\hline
Use Case & Question  &  Reasoner response    \\ \hline
Equinor  & Does the proposed BCAS302R VDS   meet the requirements for the proposed line location P-63-CW032?       & Yes      \\ \hline
Equinor  & Does the proposed product   O.M.S.SALERI S7100 meet the requirements set out in the BCAS302R VDS        & Yes      \\ \hline
Equinor  & Does an alternate VDS BMAS302R   DN80, Dafram S.p.a. valve, meet the requirements for the proposed line location P-63-CW032? & No - the BMAS302R DN80 is trunnion mounted, not floating ball valve. These types are disjoint \\ \hline
Aker BP  & Does the proposed VDS AB-GTDD00J meet the requirements for the proposed line location A-64GT0073 and PDS VD01?                 & Yes      \\ \hline
Aker BP  & Does the proposed product IKM Flux Fig.No L6RR104 meet the requirements set out in the VDS AB-GTDD00J? & Yes \\ \hline               
\end{tabular}
\caption{Reasoner questions and answers}
\label{tab:reasonerresults}
\end{table}

\begin{figure}[h!]
    \centering
    \includegraphics[width=0.8\linewidth]{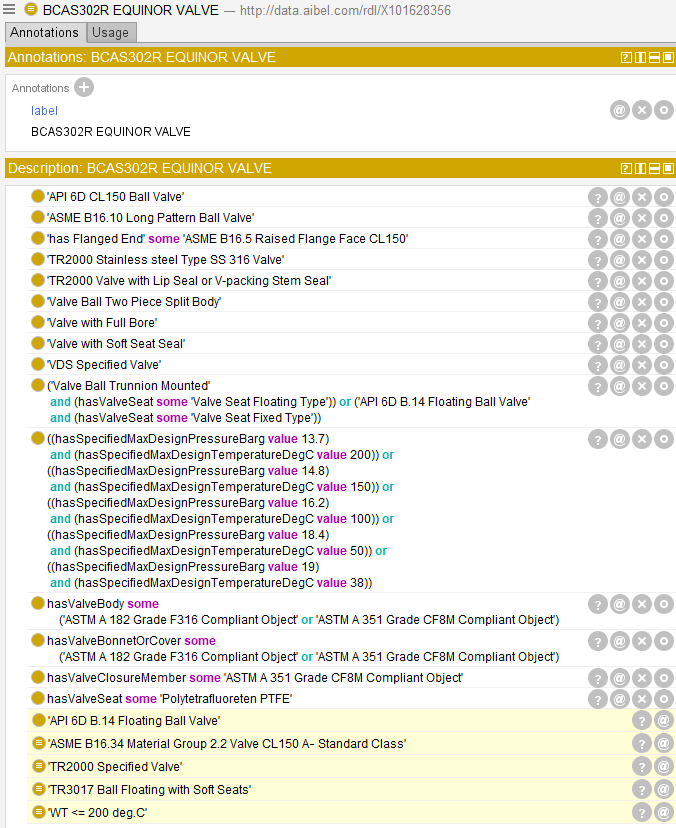}
    \caption{Evidence from the Protege Ontology Editor reasoner of the inferences made by the reasoner to ascertain that the BCAS302R -VDS meets the requirements for the proposed line location at P-63-CW032. The yellow shading shows the outputs of the reasoner.}
    \label{fig:reasoner1}
\end{figure}

\begin{figure}[h!]
    \centering
    \includegraphics[width=0.8\linewidth]{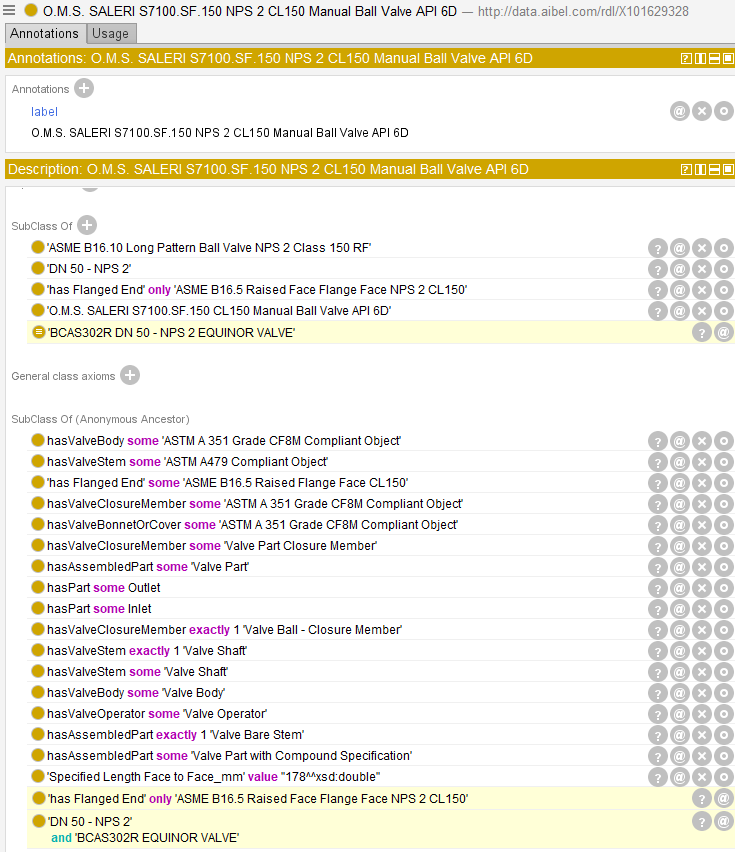}
    \caption{Evidence from the Protege Ontology Editor reasoner of the inferences made by the reasoner to ascertain that the proposed O.M.S SALERI Manual Ball Valve meets the requirements for the BCAS302R VDS. The yellow shading shows the outputs of the reasoner.}
    \label{fig:reasoner2}
\end{figure}

As examples of the performance of the reasoner we include 1) Figure \ref{fig:reasoner1}: a screen shot showing the reasoning that the BCAS302R-VDS meets the requirements for the proposed line location at P-63-CW032 and 2) Figure\ref{fig:reasoner2}: a screen shot showing that the inferences made by the reasoner to ascertain that proposed O.M.S SALERI Manual Ball Valve meets the requirements for the BCAS302R VDS. Screen shots demonstrating the outputs for the other questions in Table \ref{tab:reasonerresults} are included as supplementary materials in the GitHub repository.

\subsection{Checking completeness}

SHACL is used to check each specification for \textit{area line} and \textit{process line}. An \textit{area line} is typically an assembly of several piping items such as pipe spools, inline instruments and valves. In the ontology \textit{area line} is a subclass of \textit{PhysicalArtefact}. A process line representes the medium inside line. In this ontology \textit{ProcessLine} is a subclass of \textit{Stream}.

An area line specification must:
\begin{itemize}
    \item contain a process line, i.e., \texttt{lis:contains} relation to an instance of the class \textit{ProcessLine},
    \item reside in a plant area, i.e., \texttt{residesIn} relation to an instance of \textit{PlantArea}.
\end{itemize}.

A process line specification must include a specific list of attributes whose values must be a decimal number within an allowed range. SHACL shapes are developed that implement requirements for the area line and process line specifications. Examples of these are provided in the GitHub supplementary materials.

\section{Conclusion and recommendations} 

We have addressed goals identified in the `IEC SMART standards - from a market and industry report' \cite{IECSMART2report} specifically for 1) standards that can be interpretable by machine without human intervention and, 2) automated compliance check of data to assure conformity to standards down to individual products. For the SDOs, many of whom still offer only pdf versions of their standards, we present evidence that machine-interpretable standards are a real option for them. For ISO Smart, already on their way to digitisation, we have shown that the selection of OWL/RDF and alignment to the IDO ontology tangibly demonstrates an example of their machine-interpretable vision.

We contribute to the Body of Knowledge by demonstrating how information in engineering standards can be made machine-interpretable and usable for engineers in their work. We have developed formal, machine-readable representations for selected API and ASME engineering standards for automated reasoning and analysis. These ontology modules are conformant to a top-level ontology (the Industrial Data ISO CD 23726-3). We demonstrate machine automation of reasoning typically done by a piping engineer in VDS selection and product checking (as shown in Figures \ref{fig:reasoner1} and \ref{fig:reasoner2}).

This ontology work includes solutions to modelling issues specific to the process industry. One example being how to convert pressure-temperature (as found in ASME B16.34) to a OWL 2 machine-executable design rule. Pressure-temperature tables are used through out piping. The method used to capture this is applicable for such tables in piping class sheets, valve data sheets and in other standards like ASME B16.5.

We deliver the valve ontology as an OWL 2 ontology module available on GitHub. This module defines the set of things that are of interest to valve product designers and users and how these things can relate to one another (as shown in Figures \ref{fig:valve core} and \ref{fig:valve object properties}. 

We demonstrate the ability, through use of annotation properties, to trace back from the use of a specific class to the edition of the standard in which it is described using the \texttt{Is defined in Edition of Specification}. This is an important quality check that needs to be done in the engineering process as standards can have multiple editions and it is time consuming to cross-reference these manually. 

For industry, this is a tangible use case of the value of ontologies for a real task commonly done by very experienced (and expensive) piping engineers. The use case also shows how knowledge can be organised and shared across engineering projects and stakeholders.

The process of developing re-usable modules representing an existing shared understanding based on international standards and engineering norms supports standardisation and communication. Engineers can also reuse modules for different standards or projects without starting from scratch. This approach will help in industries in high hazard, safety critical process industries where standards compliance and knowledge sharing are critical.

We close by offering the following recommendations and areas of future work.

\begin{itemize}
    \item SDOs should accelerate their journey towards digitisation of their standards with the goal of making data tables available as ontology modules for use by industry. This requires new business models as their customers migrate from PDF-format standards to the types of ontology modules we have demonstrated here. 
    \item We recommend the adoption of IDO as a top-level ontology for the Industrial Standards community. As shown in this paper, engineering standard-based ontologies can provide reference data at an industrially relevant level of detail and enable sharing of both semantic product specification and asset information,
    \item The need for tracing references from resources in the ontology to its source in the standard document is acknowledged. To interpret content in one industry standard, the content of other standards must often be known. Industry standard ontologies must have references to resources in other industry standard-based ontologies. To achieve this ontology representation of industry standards should include annotation type references from resources within the ontology back to sections, paragraph, and cell in tables in the specific standard that are their source. 
    \item We are continuing our modelling work, focussing on other ontology modules to augment the work of engineers involved in the engineering design process and on processes to improve the infrastructure and capabilities necessary to support deployment in industry. We are also exploring the role of Generative AI to assist in the model development process. 
\end{itemize}

\section*{Acknowledgments}
The authors would like to thank DNV (Det norske Veritas) for providing access to their online ontology development tool – IOTool. We have used the OTTR Reasonable Ontology Templates and acknowledge Dr. Martin G. Skjæveland from the University of Oslo for his contribution to effective ontology development and for producing the test report to check completeness of the area line and process line specifications with SHACL and the valve specification with OWL.

\section*{Funding}
This work has been supported by DISC and the Industrial data ontology (IDO) consortium.  

DISC (DIGITALISATION – INDUSTRIALISATION – STANDARDISATION – COLLABORATION) is a formalized collaboration between Equinor, Aker BP, Aker Solutions and Aibel who all are key players in the Norwegian energy sector.

 The Industrial Data Ontology (IDO) consortium  supports and drives standardization of IDO as a top level ontology under ISO. IDO has been assigned the standard number ISO 23726-3. POSC Caesar Association leads the consortium which includes members and observers such as Aibel, AkerBP, Aker Solutions, DNV, Equinor, Grundfos, SEIIA, Siemens, Standards Norway, TotalEnergies, CFIHOS, University of Oslo, University of South Australia and University of Western Australia.

\bibliographystyle{unsrt}  
\bibliography{references}  

\end{document}